\documentclass[twocolumn,final,authoryear]{elsarticle}

\usepackage{prletters}
\usepackage{framed,multirow}
\usepackage{url}
\usepackage{amssymb} \newcommand{\stitle}[1]{\vspace{1ex}\noindent\textup{\textbf{#1}}}
\usepackage{amssymb}
\usepackage{latexsym}
\usepackage{subfigure}
\newcommand{\myNum}[1]{(\emph{#1})}
\journal{Pattern Recognition Letters}
\usepackage{arabtex}

\usepackage{epstopdf}
\newcounter{footnoteA}
\newcommand{\footnoteA}{%

  \renewcommand\thefootnoteA{\Alph{footnoteA}}
  \stepcounter{footnoteA}%

  \Footnotemark\thefootnoteA \FootnotetextA{}}

\newcommand{\myUrl}[1]{%

    \footnoteA{\url{#1}}}
\begin{document}

\vskip1pc

\begin{frontmatter}

\title{AltecOnDB: A Large-Vocabulary Arabic Online Handwriting Recognition Database}

\author[1]{Ibrahim \snm{Abdelaziz}\corref{cor1}} 
\cortext[cor1]{Corresponding author: 
  }
\ead{i.hosny@fci.cu.edu.eg}
\author[2]{Sherif \snm{Abdou}}

\address[1]{Faculty of Computers \& Information, 5 Dr. Zewail Street, Orman, 12613, Giza , Egypt}
\address[2]{Faculty of Computers \& Information, 5 Dr. Zewail Street, Orman, 12613, Giza , Egypt}

\begin{abstract}

Arabic is a semitic language characterized by a complex and rich morphology. 
The exceptional degree of ambiguity in the writing system, the rich morphology, and the highly complex word formation process of roots and patterns 
all contribute to making computational approaches to Arabic very challenging. As a result, a practical handwriting recognition system should support 
large vocabulary to provide a high coverage and use the context information for disambiguation.\\

Several research efforts have been devoted for building online Arabic handwriting recognition systems.
Most of these methods are either using their small private test data sets or a standard database with limited lexicon and coverage. 
A large scale handwriting database is an essential resource that can advance the research of online handwriting recognition.
Currently, there is no online Arabic handwriting database with large lexicon, high coverage, large number of writers and training/testing data. \\

In this paper, we introduce AltecOnDB, a large scale online Arabic handwriting database. 
AltecOnDB has 98\% coverage of all the possible PAWS of the Arabic language.  
The collected samples are complete sentences that include digits and punctuation marks. 
The collected data is available on sentence, word and character levels, hence, high-level linguistic models can be used for performance improvements.
Data is collected from more than 1000 writers with different backgrounds, genders and ages. 
Annotation and verification tools are developed to facilitate the annotation and verification phases. 
We built an elementary recognition system to test our database and show the existing difficulties when handling a 
large vocabulary and dealing with large amounts of styles variations in the collected data. 
}
\end{abstract}

\end{frontmatter}

\setarab

\section{Introduction}
\label{sec:intro}
Recently, hand-held devices such as PDAs, tablet-PCs and smartphones became very popular and are gaining a wide spread.
Recognizing natural handwriting, drawn using a finger or a stylus, is a powerful feature to have on a hand-held device
as it provides a more easy, friendly, and natural way of interaction compared to using a keypad or
keyboard to input text. As a result, there is an increased demand for a high performance on-line handwritten recognition system. 

Arabic script is written from right to left and is a strictly cursive script. Almost all letters within a word or sub-word are connected. 
The standard Arabic script contains 28 letters. Each letter has either two or four different
shapes depending on its position within the word. 
The same letter at the beginning and end of a word can have a completely different appearance. 
Along with the dots and other marks representing vowels, this makes the effective size of the alphabet about 160 characters.
Most Arabic letters contain dots in addition to the letter body, such as <^s> which consists of <s>
letter body and three dots above it. Some other letters also composed of strokes which attached to
the letter-shape body such as <k> and <.t>. These dots and strokes are called delayed strokes
since they are usually drawn last in a handwritten word-part/word. This is similar to handwriting of
Latin scripts, the cross in letters t and the dots in i and j letters, they are also usually drawn at last.
Diacritical are markings which are written either above or below a letter. The diacritics Fatha, Damma, and Kasra indicate short vowels.
Sukun indicates a syllable stop, and Fathatan indicates nunation and can accompany Fatha, Damma, or Kasra.

Arabic is a language characterized by its complex and rich morphology where a word can be composed 
of a stem plus affixes and clitics. The affixes include inflectional markers for tense, gender, and/or number. 
The clitics include some prepositions, conjunctions, determiners, possessive pronouns and pronouns. 
Arabic is also extremely ambiguous such that a word may be morphologically analyzed in more than one way \cite{soudi2007arabic}.
Disambiguation in such cases, whether by a human reader or an automatic process, is possible only by considering the context of the word. 
The morphology of Arabic poses special challenges to computational natural language processing systems. 
The exceptional degree of ambiguity in the writing system, the rich morphology, and the highly complex word formation process of roots and patterns 
all contribute to making computational approaches to Arabic very challenging. As a result, a practical handwriting recognition system should support 
large vocabulary to provide a high coverage and use the context information for disambiguation.
A publicly available corpus and enormous data samples for training and testing online handwriting
recognition systems exist in English. It has been addressed by UNIPEN, a project supported by the
US National Institute of Standards and Technology (NIST) and the Linguistic Data Consortium (LDC). 
Unfortunately, there was no reference to similar type of data for Arabic script. Several researchers, \cite{hany11,hosny2011,eraqi2011}, used to
collect their own data and use it for training and system evaluation. These private databases representing generally small dictionaries with small 
number of writers, limited lexicon or just isolated forms of letters or/and digits. Moreover, as each system is using its own data, it is hard to compare them to decide which one has a better performance. 

REGIM presented a new database called ADAB \cite{ADAB2009} for training and testing online Arabic handwriting systems.
Limitations of this database are fourfold: \myNum{i} it includes a vocabulary of 937 Tunisian town/village names which is a very small lexicon and of a limited coverage. 
\myNum{ii} The number of writers is limited (only 173). 
\myNum{iii} The number of training samples (around 20K samples) is not sufficient to train a general purpose system that handle lots of different writing styles; and
\myNum{iv} Most of the samples are isolated words and as a result, systems can not exploit the contextual knowledge for enhancing the performance.
Due to these limitations, recently several systems already achieved more than 90\% accuracy on this data. Another sentence database is proposed by \cite{elanwar07}
, However, it has a limited lexicon and the amount of data and writers is still limited.
Apparently, supporting large vocabulary, variant writing styles and context-dependant modeling impose a whole set of different challenges in a rich language like Arabic.

In conclusion, there is no robust standard database devoted for Arabic handwriting script recognition with large lexicon, 
high coverage of both letters and digits, large number of writers and a large amount 
of training/testing data. 
The availability of such a database will allow the research community to test new ideas and algorithms and tackle the real imposed challenges by the Arabic language which will help to advance 
this type of research in general.

\stitle{Our Contribution: }
In this paper, we propose Arabic Language Technology Center-Online Database (Altec-OnDB), a large vocabulary database of Arabic Online handwriting. Altec-OnDB represents a rich resource for advancing the state of the art of unconstrained 
handwriting research for Arabic and other languages such as Farsi, Persians, and Urdu-speaking who use the Arabic characters in writing, although the pronunciation is different. Altec-OnDB has the following characteristics:

\begin{itemize}
  \item It covers 98\% of the different paws, the most reliable independent writing units, of the Arabic language.  
  \item The written data is selected carefully from different available text corpora. 
  \item Altec-OnDB contains samples from 1000 different writers with different education, age and gender to allow building efficient writer-independent systems. 
  \item The collected samples are available on sentence, word and character levels. Being available on sentence level allows the application of the high-level linguistic models for performance improvements. 
  \item We propose another set, AltecOnDB Set-H , where the amount of data collected per writer is relatively high. With using this dataset, writer adaptation techniques can be employed 
    to improve the writer-independent system performance or to build a writer-dependant system.
  \item The database contains samples for digits, characters and punctuation marks. Therefore, a general purpose system can be built using our proposed database.
\end{itemize}

\stitle{Paper Overview}
Section \ref{sec:rel} describes the related work to this paper. In Section \ref{sec:acq}, we describe the data preparation and acquisition process. 
Section \ref{sec:anno} describes how we used our annotation/verification to annotate and verify the correctness of the collected data. 
Then, we list the statistics of our collected data in Section \ref{sec:stats}. In Section \ref{sec:exp} describes some preleminary  experimental results that
we got using our collected database. Finally we conclude in Section \ref{sec:conc}

\section{Related Work}
\label{sec:rel}

In this section, we present a breif review of the related work in this area of research. 
Section \ref{subsec:hwr} presents some of the related work on Arabic handwriting recognition. 
In Section \ref{subsec:db}, we review the research efforts for building handwriting databases for handwriting recognition. 
For a recent and comprehensive survey on online Arabic handwritten recognition, 
interested readers can refer to the survey in \cite{najiba2013}.

\subsection{Handwriting Recognition}
\label{subsec:hwr}
Early research on online Arabic handwriting recognition is done by 
\cite{Emami90} proposed an online Arabic handwriting recognition system based on decision-trees to recognize
words. They used a set of directional features and tested their approach using 120 post code words based on 13 Arabic letters' shapes. 

Also, \cite{Habian07} presented a structured model for recognizing online Arabic handwriting written in continuous 
form based on Hidden Markov Models (HMMs) to recognize Arabic strokes. The basic units of recognition are strokes which can be a letter or a sub-letter. 
They used data collected from six writers, each is asked to write a set of predefined words six times. 
This amounts to each writer writing 486 letters, which equals to 2916 letters written in total, which equals to 5823 strokes.

The collected data is split into two equal parts one for training and the other for testing. The reported results show 75\% accuracy on both letters and strokes evaluation.

\cite{taani08} proposed an efficient structural approach for recognizing on-line Arabic handwritten digits. The recognizer
technique is based on the identifying the changes of the slope's sign. The approach is evaluated using a collected dataset of 3000 Arabic digits collected by 100 writers. 
The reported results are within the range of 95\%.

\cite{elanwar07} proposed a system to recognize online Arabic cursive handwriting based
on Rule-based method. It perform simultaneous segmentation and recognition of word portions in an
unconstrained cursive handwritten document using dynamic programming. They used a set of geometric features based Freeman chain codes. They
evaluated their method using data collected from eight writers in total. 
Training data is composed of 317 words (1814 characters), written by four writers and the testing database is composed of 94 words (435 characters) written by other four writers.
They achieved 74\% and 95\% accuracy on word and character recognition respectively.

\cite{Izadi08} developed an SVM-based online character recognition system based on a novel feature extraction technique.
They defined relative context feature (RC) obtained from the relative
pairwise distances and the angles of the writing trajectory. They used a database proposed by \cite{Mezghani05} which contains
4896 samples and the testing set 2448 samples (which corresponds to 288 samples of each character for training and 144 samples of each character for testing).
The proposed system achieved 97\% accuracy on the testing data.

\cite{daifallah09} developed an on-line Arabic handwritten recognition system based on the new stroke segmentation
algorithm. This algorithm is based on over-segmentation method followed by segmentation enhancement, consecutive joint connections and segmentation point locating. 
The system is tested using 150 words's samples containing 720 letters. The proposed system gives a recognition rate of up to 92\% for words and 97\% for letter recognition. 

\cite{alimi97} developed an online writer-dependent system to recognize Arabic cursive words based on a neuro-fuzzy approach and genetic algorithms.
The system is tested using data collected from one writer. The training data contains 2000 characters whereas the test data is 100 replications of a single word. 
They achieved accuracy of 89\% on a character-based evaluation.

\cite{wakil89} proposed a method for the recognition of isolated handwritten Arabic characters drawn on a graphic tablet. 
Two types of features are extracted from the characters. Features that are independent of
the writer style are represented as a list of integer values, while those that are subjected to more
variations are represented using a Freeman-like chain code. The system is tested using 60 isolated characters and achieved an 89\% accuracy for character recognition.

\cite{mezghani03} proposed an online system for recognizing isolated Arabic characters. Their method is based 
is based on combining two Kohonen maps, one obtained from representing the online signal using tangents and the other using Fourier descriptors.
Using Kohonen maps topology, they could prune the error-causing nodes. The system is trained using around 5000 isolated letters samples and tested using other 2400 samples.
They reported an accuracies are  86.56\% to 93.54\% before and after pruning respectively. 
In an earlier version of this system was proposed by the same authors \cite{mezghani02}. They developed another system based Fourier descriptors representation. Recognition was
 carried out by a Kohonen memory developed using a real database. Recognition accuracy was 86\% in a database of
7344 samples of 17 classes written by 17 writers. 

\cite{biadsy06} developed a HMM-based system for recognition of Arabic handwritten words. They introduced an efficient method for handling delayed strokes 
which help in changing the handwriting signal to match the HMM expectations. For training, four writers are asked to write 800 words. To evaluate the system, 
ten writers are asked to write 280 word not in the training data. The total number of testing samples are 2358 samples. On a vocabulary of 40K words, 
they reported accuracies of 88\% and 89\% for writer independent
and writer dependant models respectively.

\cite{hany11} presented an on-line Arabic handwriting recognition using HMM (Hidden Markov Model). Delayed strokes are
removed from the on-line Arabic word to avoid the difficulty and the confusion caused by the de-
layed strokes in the recognition process. Dictionaries for all the words in the ADAB database have
been constructed with and without the delayed strokes. They used two sets of the ADAB database for training and the third one for testing and reported accuracies
within 89\%-95\%

\cite{abdelazeem11} presented an on-line handwriting recognition system for Arabic personal names based on Hidden Markov Model (HMM)
is presented. The system is trained with the ADAB-database using two different methods: manually
segmented characters and non-segmented words. This work presents a recognition system dealing
with a large vocabulary of 2800 Arabic personal names using a new lexicon reduction method that
depends on the delayed strokes formation and the number of strokes.  A test dataset of 300 handwritten personal names written by 10 writers have 
been gathered to evaluate the system performance. With dictionary reduction using their method of detecting delayed strokes, they could achieve 
an accuracy of 92\% using their test data.

\cite{eraqi2011} presented an on-line Arabic handwriting recognition based on a new on-line grapheme segmentation technique
that depends on the local writing direction. Baseline detection, delayed strokes detection, PAW (Piece
of Arabic Word) main stroke construction, and characters construction from the basic grapheme are
issues that are addressed in this paper. Experiments are performed on the ADAB-database to validate
the system and the segmentation method. They used two sets for training and a third one for testing and reported results of 87\%.
The results show a significant improvement in terms of the contribution of segmentation errors to the overall system errors while providing high performance with a simple on-line feature extraction. 

\cite{hosny2011} proposed a HMM-based system for online handwriting recognition. They employed advanced HMM training methods like 
discriminative training, writer adaptive training and context-dependant modeling. They also proposed a novel method for handling delayed strokes. 
The proposed system is evaluated using ADAB database and the reported results on the unseen data is of 97\% accuracy.

Table \ref{tab:summary} summarizes all the research efforts discussed above and state the data they use and the accuracy achieved.

\begin{table*}\scriptsize
\label{tab:summary}
\caption{\label{tab1}Summary of different works to Arabic handwriting recognition}
\centering
\begin{tabular}{|p{2.25cm}|p{5cm}|p{3cm}|p{2cm}|}
\hline
Study & Dataset & Text Type & Accuracy \\ \hline
\cite{Emami90}&120 post code words based on 13 letters'&Words &86.00\%\\ \hline
\cite{taani08}&3000 Arabic digits by 100 writers&Digits&95.00\%\\ \hline
\cite{elanwar07}&411 words by 8 writers&Words and Characters&words: 74\%, characters:92\% \\ \hline

\cite{biadsy06}&training:800 words by 4 writers, testing:2358 samples by 10 writers&Words& 88\% using 40K dictionary \\ \hline
\cite{Habian07}&2916 letters by six writers&Strokes and characters&75.00\%\\ \hline
\cite{Izadi08}&4896 training samples, 2448 testing samples&Arabic Characters&97.80\%\\ \hline
\cite{alimi97}&training: 2000 characters, testing: 100 replications of a single word&Word \& Characters&89\% character-based\%\\ \hline
\cite{wakil89}&60 isolated Arabic characters&Characters&89\% character-based\%\\ \hline
\cite{daifallah09}& 150 words (720 letters) &Words and Characters&97\% for words and 92\% for letter\\ \hline
\cite{hany11} &ADAB&Arabic Words&89\%-95\%\%\\ \hline
\cite{mezghani03}& training: 5000 isolated letters, testing: 2400 letters&Characters &86.56\%-93.54\% \\ \hline
\cite{hosny2011}&ADAB&Arabic Words&97.00\%\\ \hline
\cite{abdelazeem11} &training: 2800 name, testing: 300 names&Arabic Words (Names)&92\%\\ \hline
\cite{eraqi2011} &ADAB&Arabic Words&87\%\\ \hline

\end{tabular}
\end{table*}

\subsection{Arabic Handwriting Databases}
\label{subsec:db}
As Table \ref{tab:summary} shows, several researchers used their own data to train and evaluate their methods. Most of the used datasets
have small dictionary with limited lexicon. Moreover, none of these datasets contain samples that cover more than one word. As a result,
linguistic constraints using a higher-level language model can not be used to limit the search space improve recognition accuracy.
In addition, as most of these data is private, nobody can compare two different approaches head-to-head. 

The need to create an Arabic online handwriting database derived Research group on intelligent Machines (REGIM)
in collaboration with Institut fur Nachrichtentechnik (IfN) to create such a database.
They created ADAB database which contains 20,575 Arabic words collected from 165 different writers by writing Tunisian town names.
This dataset despite being popular and used
by many researchers, \cite{hany11,hosny2011,eraqi2011}, to validate their approaches, it has many drawbacks:
\begin{itemize}
  \item It includes a vocabulary of 937 Tunisian town/village names which is a very small lexicon and of limited coverage.  
  \item The number of writers is limited (only 173). 
  \item The number of training samples (around 20K samples) is not sufficient to train a general purpose system that handle lots of different writing styles. 
  \item Most of the samples are isolated words and as a result, systems can not exploit the contextual knowledge for enhancing the performance. 
\end{itemize}

\cite{elanwar07} proposed Online Arabic Sentence Database Handwritten on Tablet PC(OHASD). It includes 154 
paragraphs written by 48 writers of ages 24-40 years from both genders. The database contains more than 3800 words and more than 19,400 characters. 
OHASD database exist at a sentence level which make it possible to be used with contextual knowledge to improve the accuracy. 
However, still it has a limited lexicon and the amount of data and writers is limited.

Despite the existence of several individual efforts by researchers to collect handwriting datasets, most of the datasets have limited lexicon vocabulary.
Moreover, only one database, \cite{elanwar07}, offers samples on the sentence level but as shown previously, the amount of avaialble data samples is still limited.
The availability of a database that addresses these limitations is important for the research community. It will help researchers to test new ideas and target 
the real problems in recognizing the Arabic script. In this paper, we introuce a new dataset, AltecOnDB, that address all these limitations. Table \ref{tab:dbsummary} shows 
how AltecOnDB has significant coverage and offer the data at several levels (character, word and sentence) compared to exisiting databases.

\begin{table}\scriptsize
\label{tab:dbsummary}
\caption{\label{tab1}Summary of Existing Online Handwriting Databases}
\centering
\tabcolsep=0.05cm
\begin{tabular}{|p{1.6cm}|p{1.15cm}|p{1.5cm}|p{1.05cm}|p{1.5cm}|p{1.25cm}|}
\hline
Database & Words & Characters & Writers & Text Type &  Vocab. \\ \hline
ADAB&29,922&157,792&173&Words (Cities' Names)&937\\ \hline
OHASD&3,825&19,467&48&Sentences&-\\ \hline
AltecOnDB&152,680&644,530 (Segmented : 106433)&1001&Character, Word and Sentence-based&39,951\\ \hline 
\end{tabular}
\end{table}

\section{Data Preparation and Acquisition}
\label{sec:acq}
\begin{figure}[h]
  \centering
  \includegraphics[width=0.92\columnwidth]{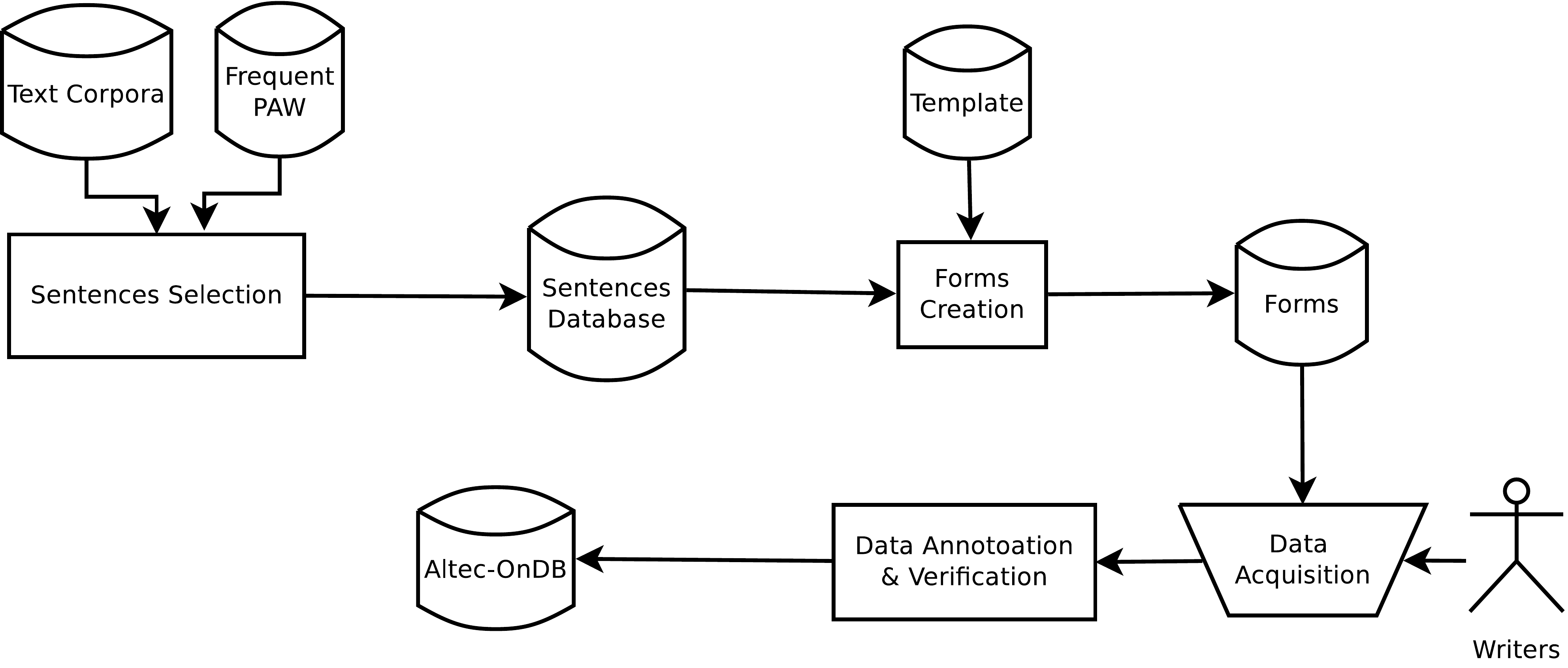}
  \caption{Block diagram of the data collection process}
  \label{fig:block}
\end{figure}

In this section, we describe the data preparation phase, the used collection device, form design and the database used to store the collected data. Figure \ref{fig:block}
shows a block diagram of the data collection process.

\subsection{Collection Device}
We used ACECAD DigiMemo L2 \cite{digimemo} device for data collection. It is a stand-alone device with storage capability that digitally captures and stores everything you write or draw with ink on ordinary paper
without the use of computer and special paper. The sheet is placed over the digitizing pad of the device and while writing, the (digital) pen’s position is recorded in the form of X-Y coordinates and 
stored in device’s on-board memory. When connected to a PC, we can access the stored forms that are written on the device where we extract and process the collected data. 
Besides the digital version of the collected data, we keep also the hard copy. The set of archived documents can be used in the future in an Arabic offline handwriting database.

\subsection{Text Corpora}
Using a corpus as the foundation of the database rather than collecting text from random sources has the advantage 
that linguistic knowledge can be automatically extracted in a more systematic and easy fashion.
Our primary goal is to select the text that cover more than 98\% of the frequent paws in the Arabic language.
In order to choose the most important paws to be covered, we analyzed our corporas by parsing the articles, segmenting it to lines, words and paws. Then, we counted the number
of occurrences for all uniquely found PAWs. The number of unique paws got from this step is about 191K. Some of these paws appeared only once, which means they are noise. 
After neglecting the noisy and the rare PAWs, this list size reduced to 36K paws; however it represents around 99\% from the all the 191k paws in frequencies.

In AltecOnDB, we used more than one corpus to enlarging our search pool for the list of sentences that satisfy our requirements. 
From the list of available sentences, we select only the sentences that meet the following requirements:
\begin{itemize}
  \item It will contribute to cover more paws. 
  \item Easy to read and write, no foreign names or expressions.
  \item Keep the original punctuation.
  \item Keep the continuation of the paragraphs inside the form as much as possible. 
\end{itemize}

We used the following text corporas:

\myNum{i} Arabic Wikipedia: produced by Linguistic Data Consortium (LDC). This is a comprehensive archive of newswire text data that has been acquired from four distinct sources of Arabic newswire which
are Agence France Presse (AFA), Al Hayat News Agency (ALH), Al Nahar News Agency (ANN) and Xinhua News Agency (XIN). There are 319 files, totaling approximately 1.1GB in compressed form (4348 MB uncompressed).

\myNum{ii} Arabic GigaWord: September 2010 dump containing around 129K articles.

\myNum{iii} Classic Arabic books: A set of very old books covering many topics and are written using the traditional Arabic language. As a result, we used only a small sample of these data as it contains some words and idioms that
 are no longer used.

In order to cover most of the Arabic PAWs, we segmented the text into paragraphs. The paragraph will be considered valuable and will be taken into account for writing if it adds value to the intended coverage; 
i.e. contains at least one uncovered paw.

\begin{figure}[t]
  \centering
  \includegraphics[width=0.8\columnwidth]{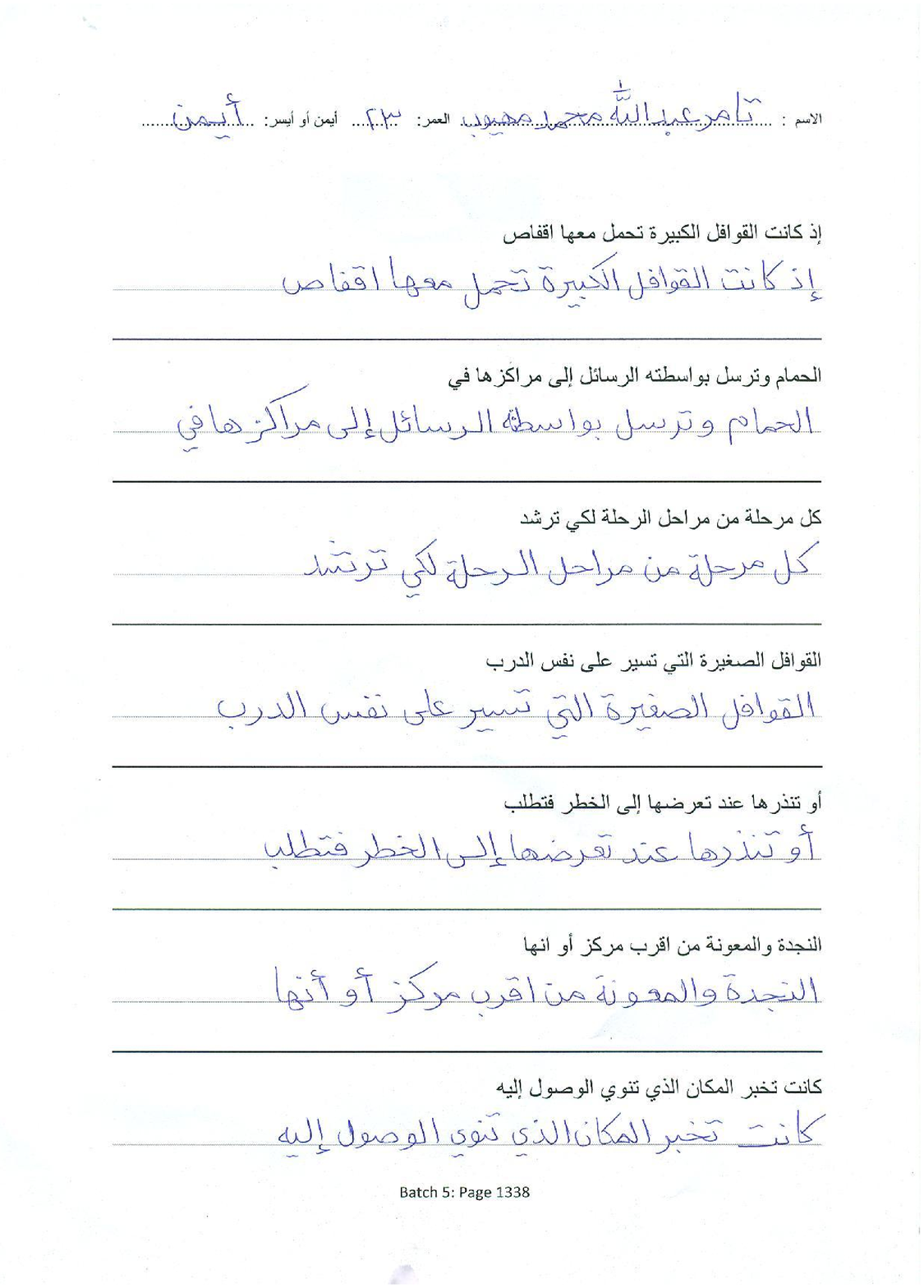}
  \caption{A sample form}
  \label{fig:form_sample}
\end{figure}
\subsection{Form Design}
\label{subsec:form}
We came up with a simple form design that does not contain too much text in terms of number of lines and number of words per line.
The first part comprises the header which contains some information about the writer like Writer Name, Age and right-handed or left-handed. 
The second part is 7 similar subparts each consists of the line text to be written (maximum 7 words per line) followed by a transparent line for the writer to write on and finally a horizontal
line separator. The last part is a footer which contains information about the document itself which are batch number, document number. These information are then used later during the annotation process.

The forms were automatically generated. A template for the form is created using Microsoft word document. After deciding on the text we will print in a form, 
the relevant text will be inserted in the template. Figure \ref{fig:form_sample} shows a sample of a form filled by one of our writers.

\subsection{Database Design}
\begin{figure}[t]
  \centering
  \includegraphics[width=0.9\columnwidth,keepaspectratio]{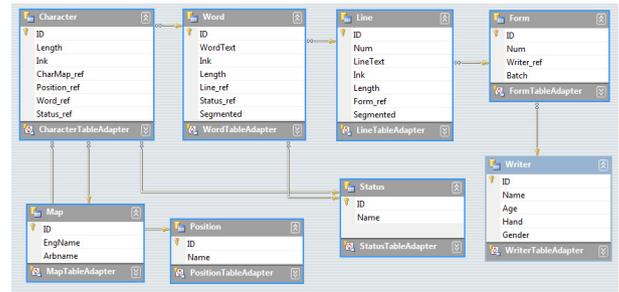}

  \caption{Database Design}
  \label{fig:erd}
\end{figure}

The data are stored as binary objects inside a MS Access database.  Figure \ref{fig:erd} shows Entity-Relationship diagram of the database where data is stored in. The Writer table contains writer information 
[name, age ,right-handed or left-handed and Gender] .The writer could write multiple forms ,So each form will be segmented into multiple lines, each line has its corresponding text and digital 
Ink which will be saved in the database as a binary object besides the line of number in the original form ,form reference that line come from and a flag to indicate whether the line has been segmented or not.
At the line level, user will segment the line into multiple words. User can tag the word as erased, unclear or not found.  Both Line, Word and Character tables have almost the same structure, the only difference is 
that they are taken from different sources; i.e. the line is taken from a form, the word is taken from a line and the character is taken from a word. A character has a position [isolated, start, middle or end] and 
map reference which states the character text.

\subsection{Data Collection process}
Once forms are ready, we give them to writers to fill in the needed information and handwrite the text in the form. Each writer is asked to fill about 7 forms. 
We tried to collect data from writers of different ages, professional backgrounds, qualifications, handedness and ages. As we show in the form design \ref{subsec:form}, we keep only
writer name, gender, age, as well as whether they were right-handed or left-handed. Even though at this stage the writer information has no significance, it could be used in future research.
Section \ref{sec:stats} show the exact statistics of the collected data and writers.  
\section{Data Annotation and Verification}
\label{sec:anno}
In this section, we describe our annotation process. We implemented a software to help us to easily acquire the data from the DigiMemo device and automatically 
segment it into lines. Then we have a set of useful tools to segment the line into words and characters. As the annotation process is a manual effort and it vulnerable to errors, we have a data
verification step in which we use another software that we implemented to make the reviewing process easier.
shows a block diagram of the data collection process.

\subsection{Data Annotation}
The annotation process represents a major challenge in constructing any handwriting database as it is so time-consuming process. An annotation application is designed to ease this process. The annotation
process go through a set of steps which we describe in the next sections.

\subsubsection{Inputting Form Information and Form Segmention}
\begin{figure}[t]
  \centering
  \includegraphics[width=0.8\columnwidth]{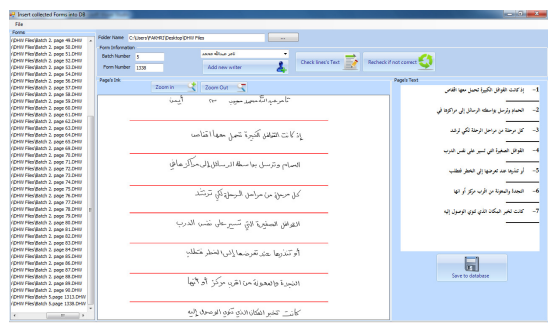}
  \caption{Lines Segmentation}
  \label{fig:linesSeg}
\end{figure}

The annotation process starts with inserting the form into the database. Figure \ref{fig:linesSeg} shows the insert page window. A user selects the folder that contains DHW files which are generated by the DigiMemo,
once the user select such folder all DHW files will be in the tree viewer. User will select what form is going to be inserted and input the form associated information which is the batch and page numbers. 
Then the associated page text is displayed on the right panel.  The user should make sure that form’s associated text matches with Digital ink. 
Then the user has to select the form writer or add a new writer if he is not in the list.
The form is segmented into lines automatically. The red lines in the form specify the lines borders. 
Once user click on Save button, lines will be inserted into the Lines table and the form information is also inserted into the Form table.

\subsubsection{Line and Word Segmentation}
The form is automatically  segmented into lines by our annotation tool. However, to make our data more useful, we segmented the lines into words and 10\% of the data is segmented into characters.
This allow researchers who target building systems for isolated characters, word and sentences recognition to use our database.

\begin{figure}[h]
  \centering
  \includegraphics[width=0.8\columnwidth]{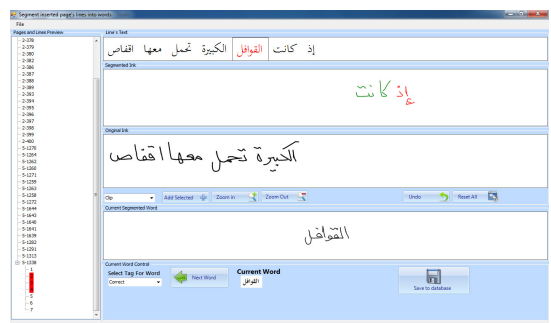}
  \caption{Word Segmentation}
  \label{fig:wordSeg}
\end{figure}

\begin{figure}[t]
 \centering             
 \subfigure[Clipping Tool]{
  	\label{fig:clip}
  	\includegraphics[width=0.3\columnwidth]{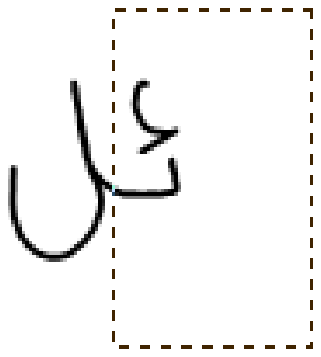}
  }
  \subfigure[Selection Tool]{
  	\label{fig:select}
  	\includegraphics[width=0.4\columnwidth]{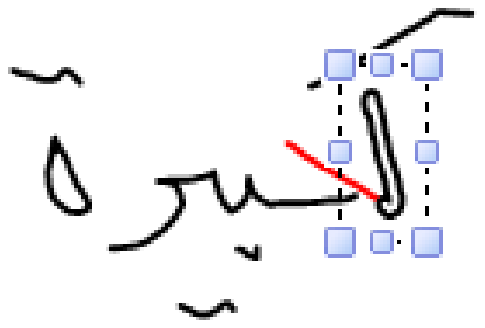}
  }
  \subfigure[Segmenter Tool]{
  	\label{fig:segment}
  	\includegraphics[width=0.68\columnwidth]{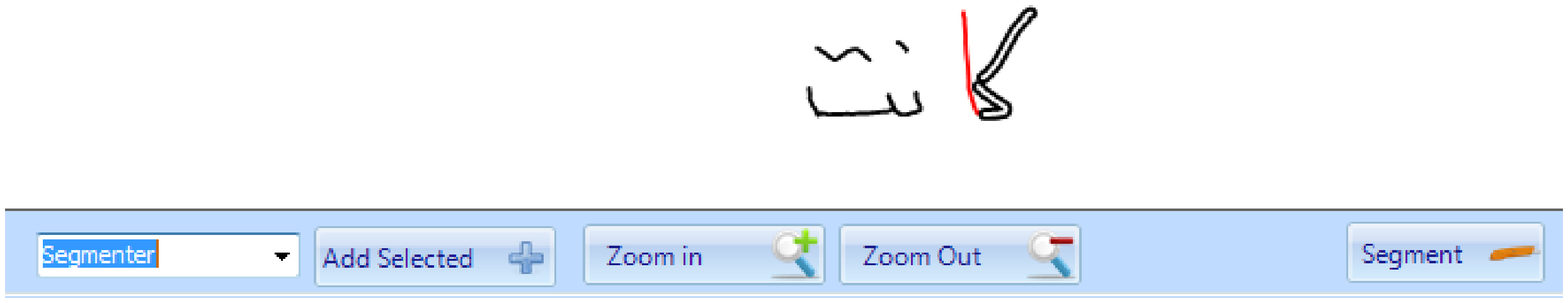}
  }
\caption{Segmentation Options.}
    \label{fig:segOptions}
\end{figure}

A set of segmentation options is implemented to ease the segmentation process. Figure \ref{fig:wordSeg} shows the lines segmentation window. The tree on the left displays all forms in the database.
After selecting a form, user can select a certain line to segment. Already segmented lines are marked as read. The upper panel display the line's text. The middle panel displays the segmented line with a color per word.
The lower panel displays the original ink. The user can use one of three segmentation tools: clipping tool, selection tool and segmenter tool. 
Clipping tool as in Figure \ref{fig:clip} allows the user to draw out a rectangle and have ink clipped inside it. It makes the segmentation an easy task whether line segmentation or in word segmentation.
In Figure \ref{fig:wordSeg}, once a word is segmented, it moved to the last panel with a possibility to tag it as correct or wrong. After the user verifies his segmented word and its correctness, he can add the segmented word which
will be placed in the segmented ink panel. 
Another segmentation tool is the Selection tool as in Figure \ref{fig:select}. It allow the user to segment the line or word through making one or more successive stroke selection. In some cases, neither the clipping nor the selection will 
be applicable, so another tool is provided. Figure \ref{fig:segment} shows an example of using the segmenter tool, in this case the selection tool is not applicable as it will select the whole stroke (both characters KAF and ALF). 
Also the clipping tool will not be able to separate the two characters as the rectangle of the clipping tool will take parts of the next character, as a result the segmenter tool will segment this stroke into two strokes, 
as a result the selection tool can select each character individually.

\begin{figure}[h]
  \centering
  \includegraphics[width=0.8\columnwidth]{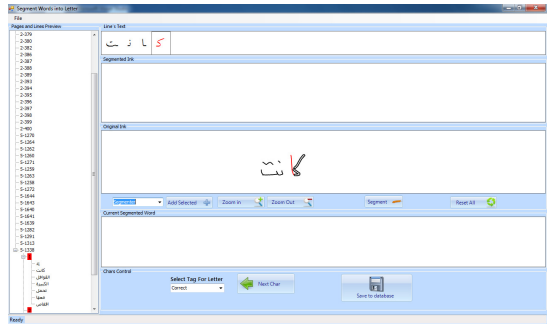}
  \caption{Character Segmentation}
  \label{fig:charSeg}
\end{figure}

Segmenting the words into characters follows the same procedure. After selecting a certain form, line and a word, it is displayed with the relevant information to allow the user to easily segment it.
Figure \ref{fig:charSeg} show the words' segmentation window. 

\subsection{Data Verification}

As most of the collection/annotation process is a manual work, it is vulnerable to errors. As a result, a verification step 
is needed to make sure that the annotated data is free of errors. We have two phases for data verification: at the word level and at the
character level. 

\stitle{Phase One: Verification at the Word Level}

\begin{figure}[t]
 \centering             
 \subfigure[]{
  	\label{fig:words_err1}
  	\includegraphics[width=0.7\columnwidth]{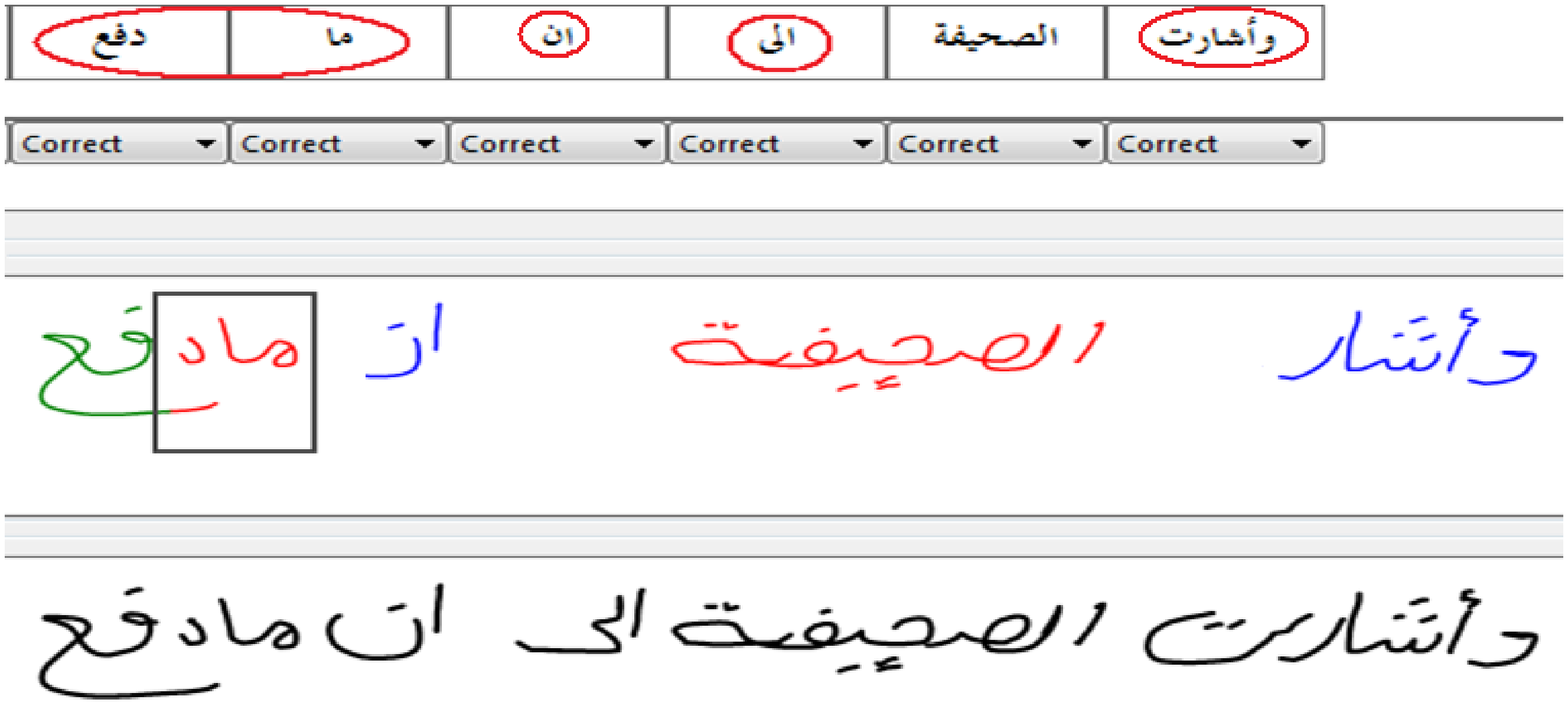}
  }
  \subfigure[]{
  	\label{fig:words_err2}
  	\includegraphics[width=0.7\columnwidth]{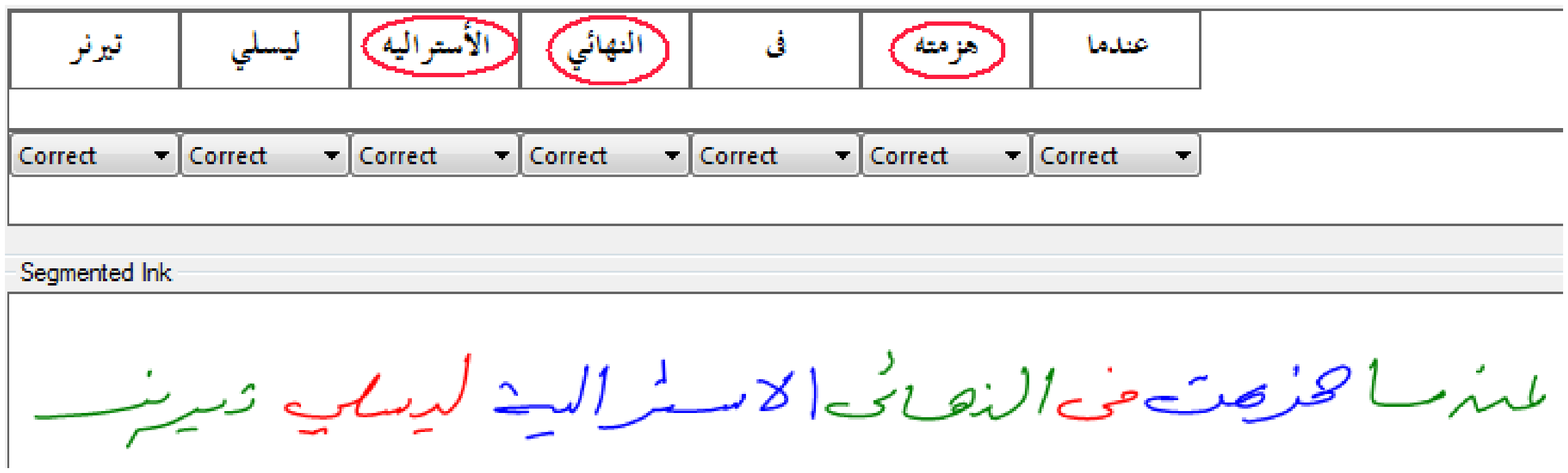}
  }
\caption{Sample Errors in Words Annotation}
\end{figure}

We found out that there are four types of possible errors:
\myNum{i} the word reference does not match the ink. \myNum{ii} A word was skipped during segmentation.
\myNum{iii} A part of the character or the word is missing and \myNum{iv} overlapping characters or dots are not well separated.
Figure \ref{fig:words_err1} shows a sample with several errors (marked with red circles). From right to left, the first error is the last character of the word is missing.
Then, two words are skipped from from the segmentation process, while the last error is an overlapping characters between two different words. Similarly,
Figure \ref{fig:words_err2} shows several cases of error iii and i. In the first case, the last letter is missed, while in the next two cases, the reference 
does not match the ink.

\begin{figure}[h]
  \centering
  \includegraphics[width=0.8\columnwidth]{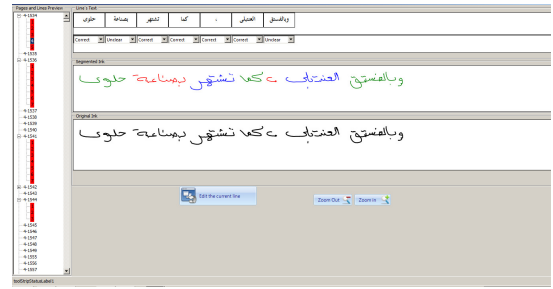}
  \caption{Word Annotation Review}
  \label{fig:lineRev}
\end{figure}

Similar to the segmentation process, we developed an effective tool to facilitate the reviewing process. 
Figure \ref{fig:lineRev} shows the window of our tool that is used for reviewing words annotation.
The user can pick a certain page and certain line, then we display the original and the segmented data differentiated by different 
colors. Under each word, there is a possibility to change the status of that word from correct to unclear or wrong. The user
can even change the text associated with the word itself.

\stitle{Phase Two: Verification at the Character Level}

\begin{figure}[t]
 \centering             
 \subfigure[]{
  	\label{fig:chars_err1}
  	\includegraphics[width=0.45\columnwidth,height=3cm]{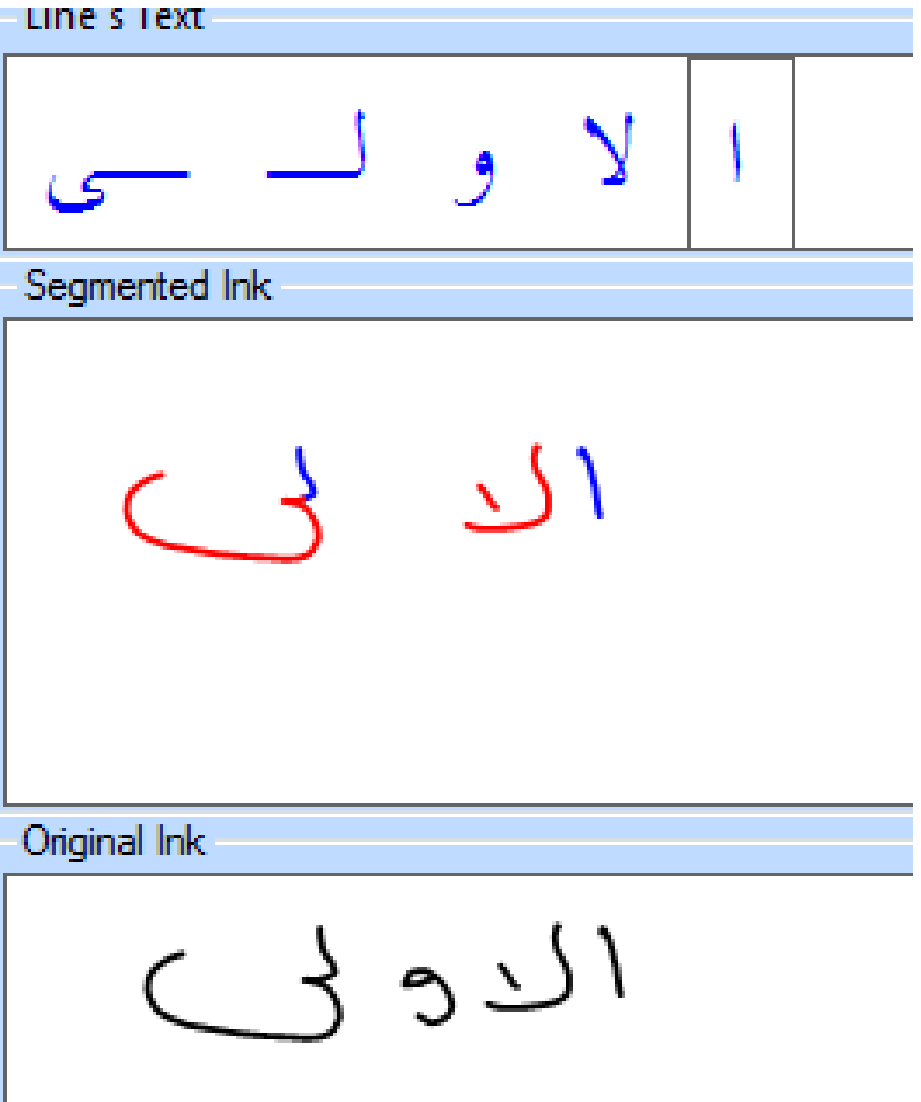}
  }
  \subfigure[]{
  	\label{fig:chars_err2}
  	\includegraphics[width=0.45\columnwidth,height=3cm]{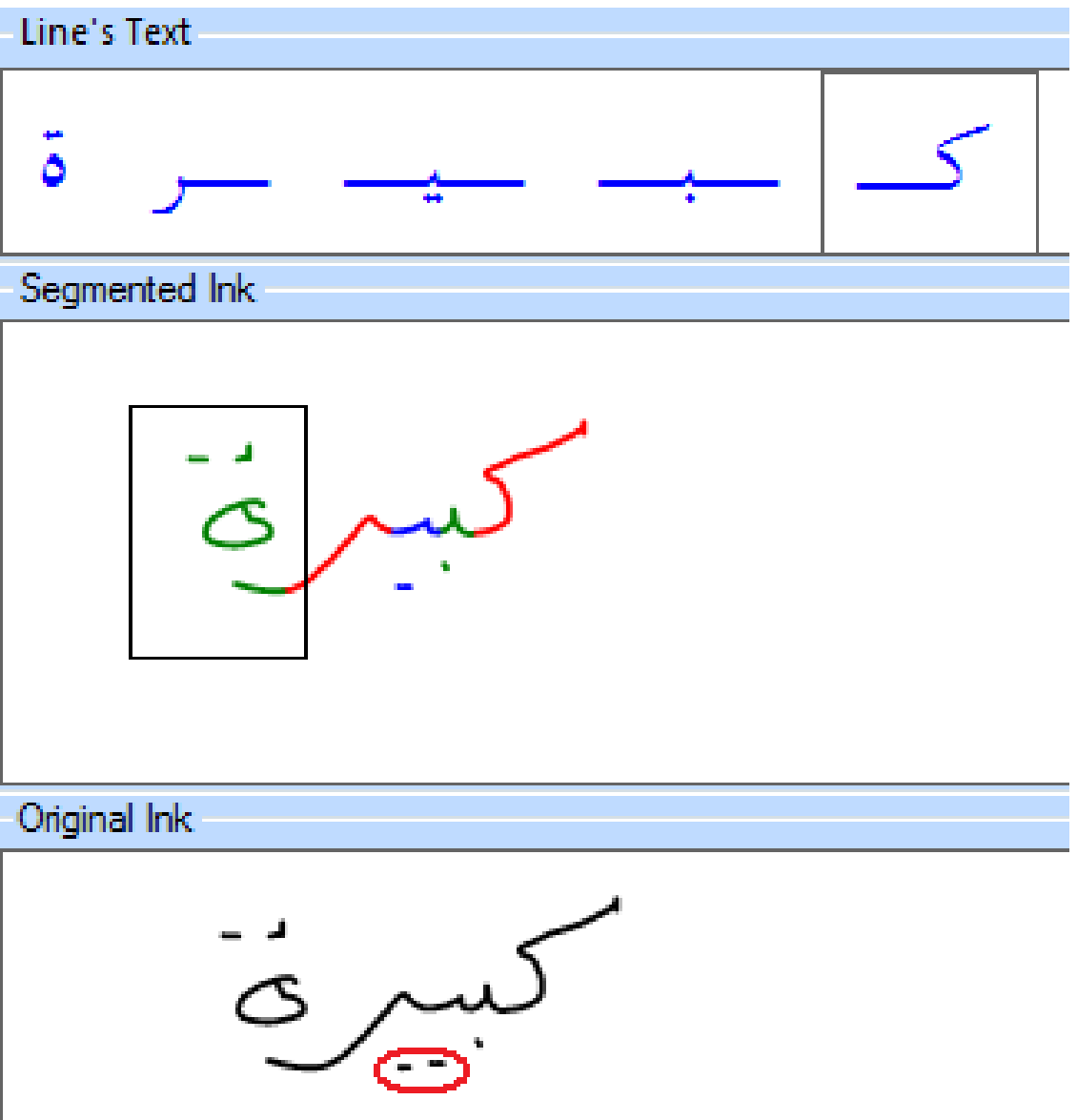}
  }
\caption{Sample Errors in Characters Annotation}
\end{figure}

There are a set of possible errors during character segmentation: 
\myNum{i} Changing the character tag from correct to unclear and vice versa.
\myNum{ii} Overlapping characters and \myNum{iii} a character or part of it is missed. Figure \ref{fig:chars_err1}
shows a sample where a character is missing, while Figure \ref{fig:chars_err1} shows a sample of overlapped characters that
are not well separated and a missing dot for the third character.

\begin{figure}[h]
  \centering
  \includegraphics[width=0.8\columnwidth]{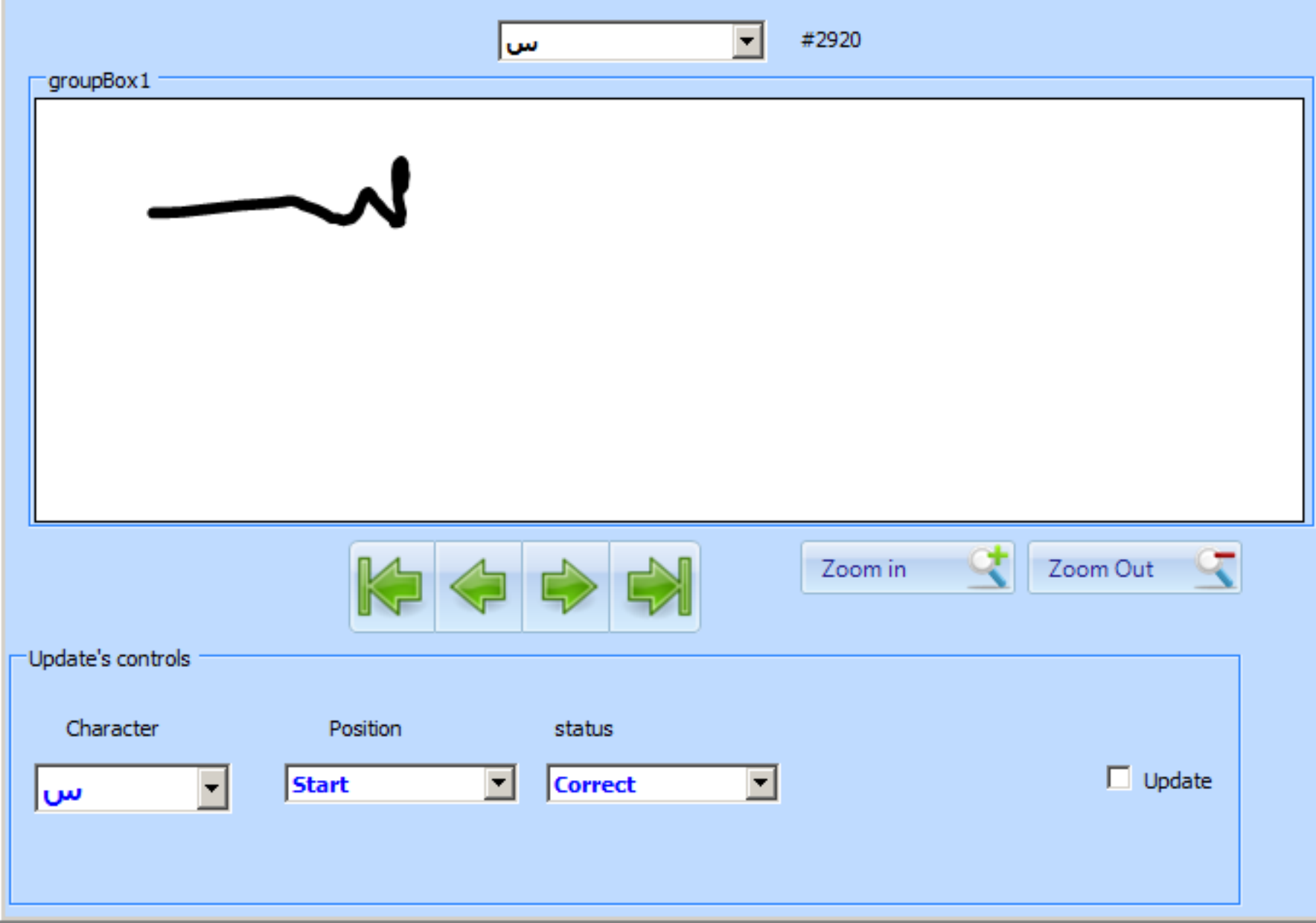}
  \caption{Character Annotation Review}
  \label{fig:charRev}
\end{figure}

Figure \ref{fig:charRev} shows the character annotation review window. The user selects which character he wants to review, 
then the appropriate samples are loaded. Then, he can use the navigation arrows to navigate between different samples. In the lower 
part of this window, we display the sample information like the character, position and correctness. These information can be updated
by the user, if he found a certain annotation error. Similarly,

\section{Database Statistics}
\label{sec:stats}

In this section, we list some statistics about the collected data. 

\subsection{Writers Distribution}
\begin{figure}
 \centering             
 \subfigure[Gender Distribution]{
  	\label{fig:gender}
  	\includegraphics[width=0.4\columnwidth]{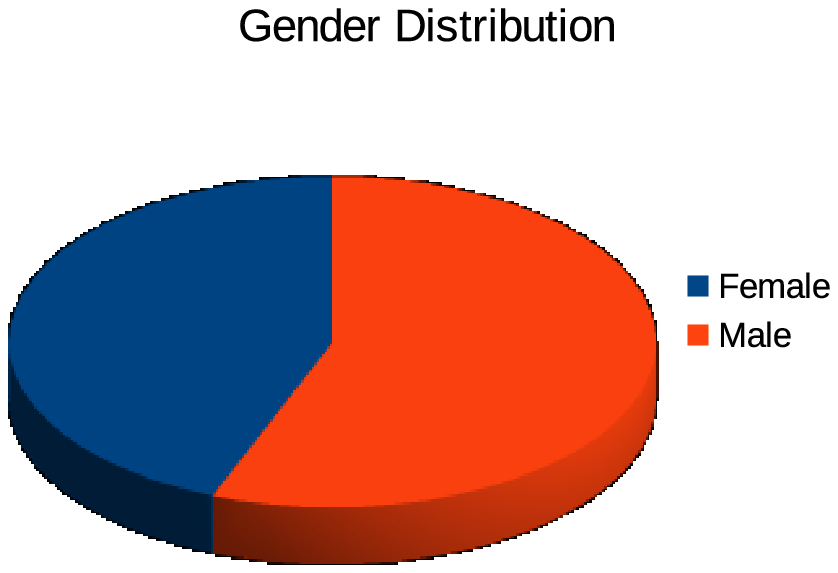}
  }
  \subfigure[Age Distribution]{
  	\label{fig:age}
  	\includegraphics[width=0.4\columnwidth]{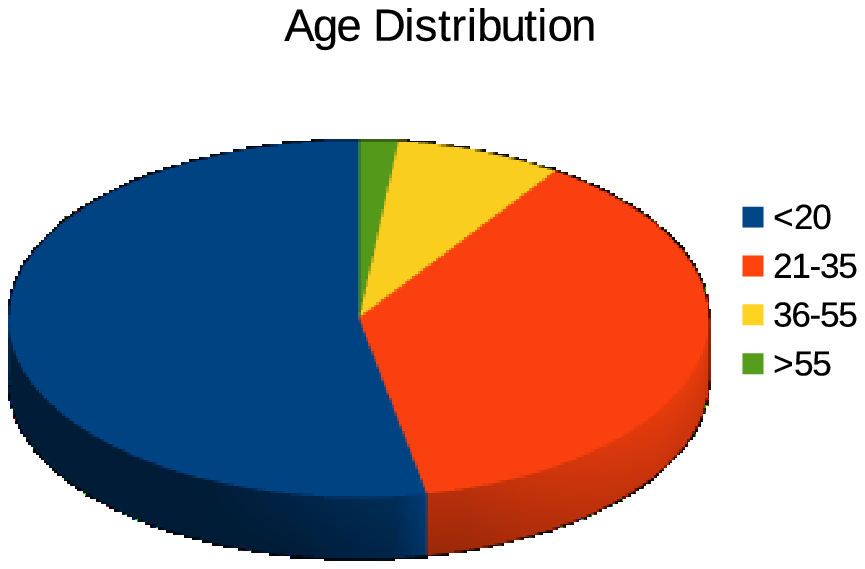}
  }
\caption{Segmentation Options.}
    \label{fig:distrn}
\end{figure}

We tried to select our writers in a way such that there are samples from different ages, sexes and education levels.
In Figure \ref{fig:distrn}, we show the writers distribution. Figure \ref{fig:gender} shows the percentage of males writers vs. females. As we can,
we have almost equal coverage for both genders. Similarly, Figure \ref{fig:age} shows the distribution of the writer ages. Most
of the writers we have are under 20 years as in our initial phase, we were targeting high schools and universities. 
The next range is writers with less than 35 years old, then few percentages for older writers. 

\subsection{Data Statistics}

Each writer is asked to fill about 7 forms. We collected handwriting samples from 1001 different writers comprised of 
men and women from various professional backgrounds, qualifications, and ages. 
We were interested in keeping track of writer’s gender, age and whether they were right-handed or left-handed. 
Even though at this stage the writer information has no significance, it could be used in future research.

We have collected 4512 different forms containing 31124 lines of handwritten text all together. 
A total of 1526580 word instances representing a vocabulary of 39951 unique words occur in the database. 
A total of 947184 PAWs instances representing 15655 unique PAWs. Detailed statistics are shown in Table \ref{tab:stats}

\begin{table}\scriptsize
\centering
\caption{Database Statistics }
\tabcolsep=0.10cm
\begin{tabular}{|l|r|}
 \hline
     & \multicolumn{1}{c|}{Count}  \\ \hline
    Writers & 1001 (M: 561, F:440)\\ \hline
    Pages & 4512 \\ \hline
    Avg. Pages/Writer & 4 \\ \hline
    Lines &31124 \\ \hline
    Segmented Words & 152680 \\ \hline
    Segmented Characters& 106433 \\ \hline
    PAWs & 325508 \\ \hline
\end{tabular}
\label{tab:stats}
\end{table}
\subsubsection{AltecOnDB Set-H: Larger Amount of Data per Writer}
Table \ref{tab:stats} shows the statistics of the whole database. As we can see, the amount of data written per writer is relatively small (4 pages, 196 words). This amount of data per writer 
is not enough if somebody wants to utilize the writer adaptation techniques to build a stronger recognition system or to adapt the system to a certain writer. 
To overcome this limitation, we collected another set of data called Set-H. 
This set is collected by 16 writers where each writer is asked to write 11 pages with average 750 words. Compared to the average amount of data per writer shown above, this amount is relatively high.
Table \ref{tab:setH} shows the detailed statistics about Set-H.

\begin{table}\scriptsize
\centering
\caption{AltecOnDB Set-H Statistics }
\tabcolsep=0.10cm
\begin{tabular}{|l|r|}
 \hline
    & Count  \\ \hline
    Number of writers & 16\\ \hline
    Number of pages & 176 \\ \hline
    Number of lines & 1717\\ \hline
    Number of Words & 12853 \\ \hline
    Avg. Pages/Writer & 11 \\ \hline
    Avg. Words/Writer & 750 \\ \hline
\end{tabular}
\label{tab:setH}
\end{table}

\subsection{Character Distribution}
\begin{figure}
 \centering             
 \subfigure[Arabic Letters Frequency Analysis]{
  	\label{fig:arAnalysis}
  	\includegraphics[natwidth=1129, natheight=340,width=1.0\columnwidth]{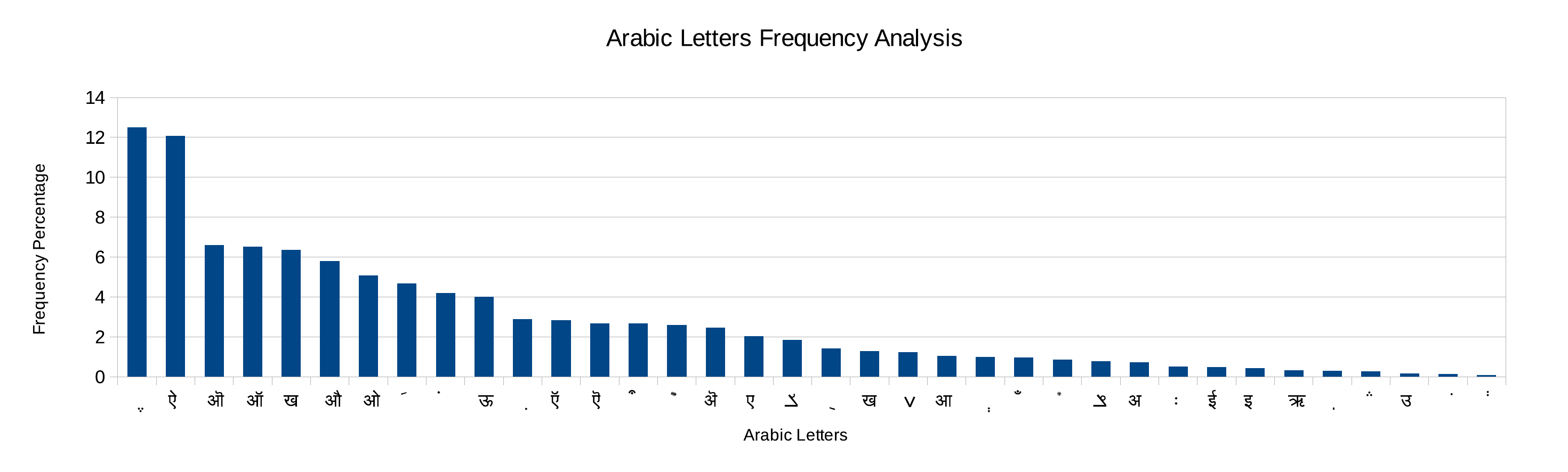}
  }
  \subfigure[AltecOnDB Segmented Characters (Position-Independent) Coverage]{
  	\label{fig:altecSegChar}
  	\includegraphics[natwidth=1104, natheight=340,width=1.0\columnwidth]{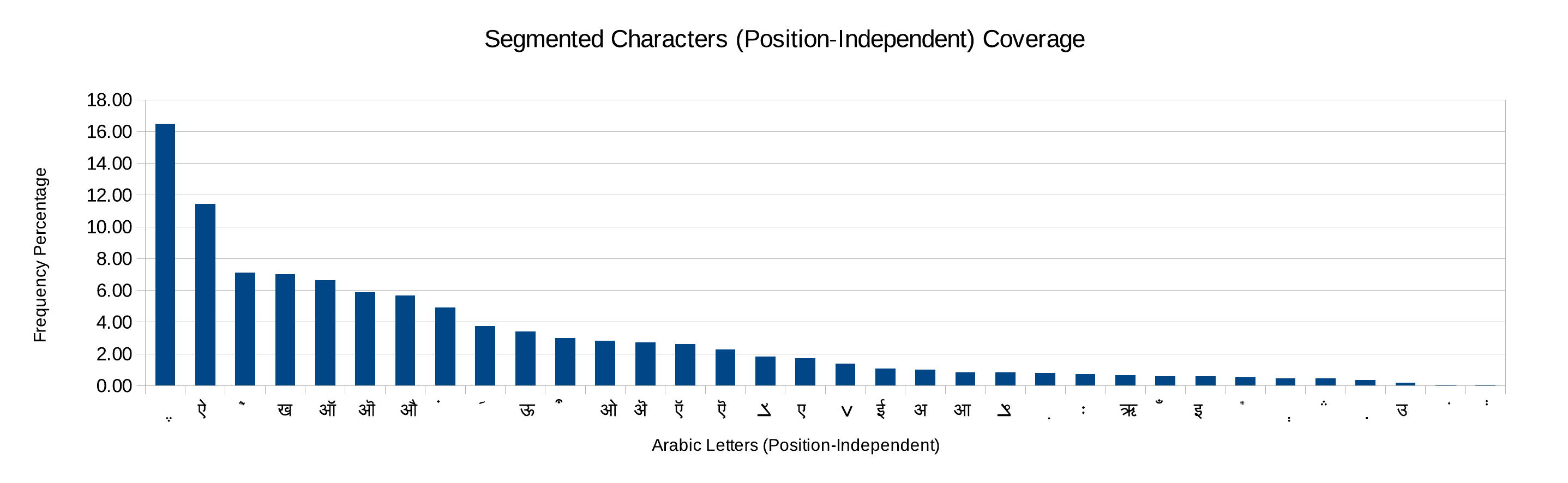}
  }
\caption{Arabic Letters Distribution: Arabic Language vs. AltecOnDB Coverage.}
    \label{fig:distrn}
\end{figure}

As stated earlier, we segmented only 10\% of the data at the character level. In this section, we show that the we collected a representative amount of data for each character
according to its frequency in the Arabic language. In a recent work by \cite{AnalysisArabic}, they study the Arabic letters frequency distribution using text sources that  add up to 3,378 pages, 
generating 1,297,259 words, or, 5,122,132 letters. The letter frequency distribution for the data they consider is shown in Figure \ref{fig:arAnalysis}. As we can see, some letters are very 
frequent than others especially the letters that exist in most of the Arabic words like <A> and  <l>. On the other hand, letters such as <.z> and  <'> are infrequent according to the conducted study.

Figure \ref{fig:altecSegChar} shows the percentage of letters occurrences in the 10\% amount of the data that we segmented. The total number of characters segmented sum up to 108133 samples.
As the Figure shows, the representation of each character correspond to its frequency in the Arabic language (see Figure \ref{fig:arAnalysis}). This shows that the text corpus used in the data collection 
truly represent the Arabic language to a high extent. Moreover, one can use our data of the segmented characters and build an effective Arabic handwritten character recognition system. 

\section{Experimental Results}
\label{sec:exp}

In this section, we report a number of experiments on the database with an elementary HMM-based recognizer. 
The conducted experiments show how difficult the recognition of the large-vocabulary Arabic handwriting samples is. Moreover, it also show 
that due to the large number of writers and the variations in their styles, having a generic model that capture all the existing variabilities is not 
a straight forward task. In Section \ref{subsec:rec}, we describe the details of the recognition system used in the evaluation. 
Section \ref{subsec:adabEval} shows that it is possible to build a generic handwriting 
recognition system using our database and test it on other databases collected by a different device.
Section \ref{subsec:diffVocab} shows a set of experiments 
conducted using AltecOnDB Set-H as a testing set while varying the size of the dictionary supported. 
Finally, we show in Section \ref{subsec:lm} that utilizing the availability of data at the sentence level and employing a higher 
level linguistic model can significantly improve the recognition results.

All experiments are run using a Lenovo z560 machine with 4G RAM and 2.53GHz Intel core i5 processor. The machine is running 64-bit Microsoft Windows 7.

\subsection{Recognizer Description}
\label{subsec:rec}
In this paper, we used a Hidden Markov Models (HMM)-based system. We briefly describe below how the system is built.

\stitle{Preprocessing}
The goals of the preprocessing phase are: reduce/remove imperfections caused by acquisition devices, smooth the
irregularity generated by inexperienced writers having an erratic handwriting and minimize handwriting variations irrelevant for 
pattern classification which may exist in the acquired data. Due to the variation in writing speed, the acquired points are not distributed evenly along
the stroke trajectory. We re-sample the acquired points to get a sequence of points which is equidistant.
Finally, we remove the hooks that may appear with sensitive pens at the beginning or end of the strokes due to inaccuracies in rapid pen-down/up
detection and erratic hand-motion.

\stitle{Feature Extraction}
We use a simple set of features that are measured per point to build our recognizer. A 32-directional chain code is used for boundary description. 
Curliness is a feature that describes the deviation from a straight line in the vicinity of (x(t), y(t)). Aspect ratio characterizes the height-to-width 
ratio of the bounding box. Curvature of a curve at a point is a measure of how sensitive its tangent line is to moving that point to other nearby points.

\stitle{Recognizer}
The proposed system is based on Hidden Markov Models (HMM). The HMM is a finite set of states, each of
which is associated with a (generally multidimensional) probability distribution. Transitions among the states
are governed by a set of probabilities called transition probabilities. In our system, we use left to right HMM model with
different number of states per model according to how complex the model shape is.

Arabic contains 28 different letters, but as these letters are position dependent it will map to 103 different
shapes. In our proposed system, we have 115 different models. These models include Arabic letters with their
different shapes (103 models), 10 English digits (0-9), Arabic MAD symbol (~) and English Capital V letter.

We build a mono-grapheme system which is based on 115 different models (position-dependent) using the Maximum Likelihood
(ML) training to maximize the probability of the training samples generated by the model. Although we can build more advanced models (\cite{hosny2011}) which will
significantly give better performance, we want to only show the applicability of the database and the difficulties inherited when working with large vocabulary 
and huge number of handwriting styles.

\subsection{Training Data}
To build our elementary recognizer, we did not use the full AltecOnDB data, rather we used only 30\% of the data. As we will see later in the evaluation part, 
although we did not use the full amount of data, the recognizer still produce good results. Table \ref{tab:training} shows the detailed statistics about the data used in the training 
phase. 

\begin{table}\scriptsize
\centering
\caption{Training Data Statistics (30\% of AltecOnDB)}
\tabcolsep=0.10cm
\begin{tabular}{|l|r|}
 \hline
    & Count  \\ \hline
    Number of writers & 279\\ \hline
    Number of Segmented Characters & 21201\\ \hline
    Number of Words & 42967 \\ \hline
    Unique Words & 15004 \\ \hline
\end{tabular}
\label{tab:training}
\end{table}

\subsection{Experiment 1: Evaluation on ADAB database}
\label{subsec:adabEval}

ADAB database is developed in cooperation between the Institut fuer Nachrichtentech-nik (IfN) and the Research group on Intelligent Machines (REGIM). 
The database consists around 20K samples written by more than 170 different writers, most of them selected from the narrower range of the National school of Engineering of Sfax (ENIS). 
The database is divided to 4 sets. Details about the number of files, words, characters, and writers for each set 1 to 4 are shown in Table \ref{tab:adabStat}.
None of this data is included in the training data of our recognizer and we use the four sets for evaluation.

\begin{table}\scriptsize
\centering
\caption{ADAB Database Characteristics}
\label{tab:adabStat}      
\begin{tabular}{|l|l|l|l|l|}
\hline
Set	& Files	& Words	& Characters	& Writers \\\hline
1 & 5037 & 7670 & 40500 & 56 \\ \hline
2 & 5090 & 7851 & 41515 & 37 \\  \hline
3 & 5031 & 7730 & 40544 & 39 \\   \hline
4 & 4417 & 6671 & 35253 & 41 \\  \hline
\end{tabular}
\end{table}

The evaluation results of our system using ADAB database are shown in Table \ref{adabresults}. We report the accuracies for Top 1,5 and 10 results, respectively.
As Table \ref{adabresults} shows, although AltecOnDB is collected using a different device than what have been used
for collecting ADAB database, our elementary recognition system acheives good results on all testing sets. As we can see from the high accuracy of Top 5 and 10, the correct result of most of the samples is 
included as one of the top 5 or 10 possible solutions. This means that with few optimizations on the preprocessing or the modeling techniques, the system performance can be improved significantly.

\begin{center}
\begin{table}\scriptsize
\centering
\caption{ADAB Evaluation Results (Accuracy) }
\label{adabresults}      
\begin{tabular}{|l|l|l|l|}
\hline
Set	& Top 1	& Top 5	& Top 10\\     \hline
Set 1	& 83.15 & 92.44 &93.94 \\ \hline
Set 2	& 83.08 & 92.70 & 94.47\\ \hline
Set 3	& 86.98 & 95.02 &96.21 \\  \hline
Set 4	& 87.75 & 95.47 & 96.59\\          \hline
\end{tabular}
\end{table}
\end{center}

\subsection{Experiment 2: Varying Vocabulary Size}
\label{subsec:diffVocab}

In these set of experiments, we use AltecOnDB Set-H (See Table \ref{tab:setH}) as our testing set. We show the effect of increasing the vocabulary size on the recognition system. As the vocabulary size 
increases, the search space of the recognition system becomes larger which degrades the performance in terms of both speed and accuracy. 
The Arabic language has large lexicons containing 30,000 to 90,000 words \cite{wshah2010novel}. 
Therefore, an effective recognition system for a language like Arabic should support a large vocabulary which 
of course impose several difficulties.
Table \ref{hresults} show the recognition accuracies of our elementary recognizer while varying the supported
dictionary size. We vary the dictionary size from 5000 to 64000 unique Arabic Words. The relatively low recognition 
accuracy of can be attributed to various reasons: 
\myNum{i} We do not use a language model which can improve the results significantly (see Section \ref{subsec:lm}).
\myNum{ii} The database contains real, natural Arabic handwriting from large number of handwriting styles variations. This makes building an effective recognition system that could handle
these variabilities is a challenging task. We are not aware of any recognition rates of similar data to compare with.

\begin{center}
\begin{table}\scriptsize
\centering
\caption{ADAB Evaluation Results (Accuracy)}
\label{hresults}      
\begin{tabular}{|l|l|l|l|l|}
\hline
      & \multicolumn{4}{c|}{Dictionary Size}  \\ \hline
	& 5000& 10000& 20000& 64000\\     \hline
Accuracy& 66.83 & 64.26 & 61.06 & 34.30 \\ \hline
\end{tabular}
\end{table}
\end{center}

\subsection{Experiment 3: Using Higher-Level Language Model}
\label{subsec:lm}

In this experiment, we argue that using a higher level language model with the models that we built using the training data can acheive higher performance than without using the linguistic information.
We use two passes for evaluation. In the first pass, we use the most
discriminant and computationally affordable knowledge sources which are
mono-grapheme HMM model with a bi-gram language model. The output
of the first pass is a word lattice which represents a search space with
reduced sets of hypotheses. The lattice includes several alternative words
that were recognized at any given time during the search. It also typically
contains other information such as the time segmentations for these words,
and their HMM and language scores. In the second pass, we rescore this lattice
with more powerful and expensive knowledge sources which are mono-grapheme HMM model and a fifth-gram language model. 
To build this language model, we used a text corpus collected from crawling Aljazeera news website \cite{aljazeera}. We collected around 700 MB
of text containing 132 million words, each word is four characters on average. The language model is built using SRI language modeling toolkit \cite{srilm} with its default parameters. The lattice error
rate is typically much lower than the word error rate of the single best hypotheses produced for each sentence. The multi-pass systems implementation
is a successful approach to break the tie between speed and accuracy. With this approach, it is possible to improve decoding accuracy with minor degradation in decoding speed.

Table \ref{lmresults} validates shows the results of running the two passes on AltecOnDB Set-H using a 64,000 dictionary. The difference is that we used the sentence level of Set-H rather than the separated words. 
Comparing the results with those obtained in Table \ref{hresults}, we can see that both phases acheived better performance. In Pass one, we used a bi-grams language model and a six-grams language 
model is used in pass two.

\begin{center}
\begin{table}\scriptsize
\centering
\caption{AltecOnDB Set-H: Using Higher-Level Language Model (Accuracy) with 64K vocabulary}
\label{lmresults}      
\begin{tabular}{|l|l|l|}
\hline
Set	& Pass One	& Pass Two	\\     \hline
Set-H	& 34.30 & 51.65  \\ \hline
\end{tabular}
\end{table}
\end{center}
\section{Conclusion and Availablity}
\label{sec:conc}
In this paper we introduced AltecOnDB, a large scale online Arabic handwriting database. 
In contrast to the currently available Arabic databases, our database: \myNum{i} includes high coverage for all 
the possible units of the Arabic language. \myNum{ii} Data is collected from large number of writers with different ages and backgrounds; and \myNum{iii} the collected samples are available at the sentence, word and character levels, therofore it allows the application of the high level linguistic models for performance improvments. 
The database is available online (\cite{altec}) and can be downloaded by requesting a copy from ALTEC.

\section{Acknowledgment}
\label{sec:ack}
We would like to acknowledge the Arabic language Technologies Center (ALTEC) for sponsoring the 
efforts associated with collecting and building our online handwriting database. 
Also special thanks for the Acknowledgment (ITIDA) for their initiative to fund such activities.

\bibliographystyle{model2-names}

\end{document}